%% file: paper.tex
\documentclass[10pt,twocolumn,letterpaper]{article}

\usepackage{wacv}
\usepackage{times}
\usepackage{epsfig}
\usepackage{graphicx}
\usepackage{amsmath}
\usepackage{amssymb}

\def\wacvPaperID{1056} 

\wacvfinalcopy 

\ifwacvfinal
\def\assignedStartPage{1} 
\fi

\input{tr-commands}

\ifwacvfinal
\usepackage[breaklinks=true,bookmarks=false]{hyperref}
\else
\usepackage[pagebackref=true,breaklinks=true,colorlinks,bookmarks=false]{hyperref}
\fi

\ifwacvfinal
\else
\fi

\newcommand{\numberOfNodes}{n_\mathrm n}
\newcommand{\numberOfKeyframes}{n_\mathrm k}
\newcommand{\nodeDistance}{d}
\newcommand{\nodePosition}{\mathbf p}
\newcommand{\nodeOrientation}{\mathbf \omega}
\newcommand{\action}{\mathbf a}
\newcommand{\shape}{\mathbf s}
\newcommand{\motion}{\mathbf m}
\newcommand{\goal}{\mathbf g}
\newcommand{\actions}{\mathcal A}
\newcommand{\shapes}{\mathcal S}
\newcommand{\motions}{\mathcal M}
\newcommand{\goals}{\mathcal G}
\newcommand{\absoluteMotion}{\motion_\mathrm a}
\newcommand{\relativeMotion}{\motion_\mathrm r}
\newcommand{\latentSpace}{\mathcal Z}
\newcommand{\latentCoordinate}{\mathbf z}
\newcommand{\dataDistribution}{\mathcal Y}
\newcommand{\dataSample}{\mathbf y}

\newcommand{\generator}{G}
\newcommand{\discriminator}{D}
\newcommand{\learnableParameters}{\theta}

\begin{document}

\title{Deep Generative Modelling of Human Reach-and-Place Action}

\author{Connor Daly\quad Yuzuko Nakamura\quad Tobias Ritschel\\
University College London\\
{\tt\small zcbecda@ucl.ac.uk\quad y.nakamura@ucl.ac.uk\quad t.ritschel@ucl.ac.uk
}
}

\maketitle

\begin{abstract}
The motion of picking up and placing an object in 3D space is full of subtle detail.
Typically these motions are formed from the same constraints, optimizing for swiftness, energy efficiency, as well as physiological limits.
Yet, even for identical goals, the motion realized is always subject to natural variation.
To capture these aspects computationally, we suggest a deep generative model for human \emph{reach-and-place} action, conditioned on a start and end position.
\\
We have captured a dataset of 600 such human 3D actions, to sample the $2\times3$-D space of 3D source and targets.
While temporal variation is often modeled with complex learning machinery like recurrent neural networks or networks with memory or attention, we here demonstrate a much simpler approach that is convolutional in time and makes use of (periodic) temporal encoding.
Provided a latent code and conditioned on start and end position, the model generates a complete 3D character motion in linear time as a sequence of convolutions.
Our evaluation includes several ablations, analysis of generative diversity and applications.

\end{abstract}

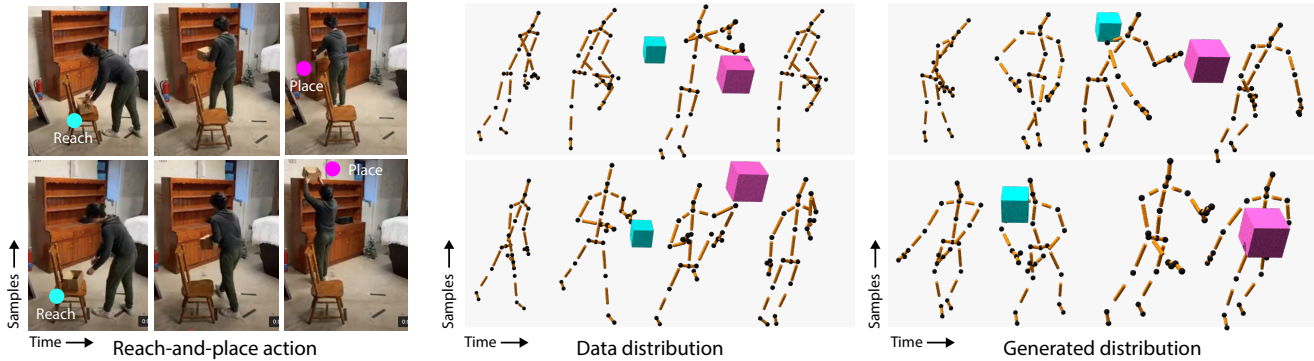
\mycfigure{Teaser}{We capture human subjects performing a reach-and-place action over time (horizontal) for several configurations of goals (rows), denoted pink and cyan.
A video of the action is shown \textbf{left}.
Off-the-shelf 3D capture produces samples of a distribution of motions, conditioned on goals \textbf{(middle)}.
We propose a neural network to generate samples from this goal-conditioned distribution of motions \textbf{(right)}.
Data distribution samples are subject to spatial noise, not present in generated samples.
Samples and goals shown are random.}

\mysection{Introduction}{Introduction}
Individuals perform complex motions with the motivation of certain goals to reach or meet, energy efficiency, swiftness, as well as physiological constraints.
Yet, even for identical tasks, the resulting motion is subject to natural variation.
Modelling all kinetic, physiological and neural relations is an elusive goal.
We suggest a method to capture all these aspects computationally, based on machine learning.

We use a deep generative model, that, provided with a start and and end goal of a motion, produces the in-between animation.
Instead of giving one answer ---or even worse, an average of all possible answers--- it captures natural variation by enabling sampling, \ie producing an infinite stream of possible motions samples, from the conditional distribution of all motions that fulfill the goals.
This model is trained adversarially \cite{goodfellow2014generative}, supervised on raw scanned 3D motion trajectories and the start and end goals.

We contribute a dataset of 600 human goal-driven 3D motions, to sample the $2\times3$-D space of 3D source and targets.
While temporal variation is often modeled with complex learning machinery like recurrent neural networks or networks with memory or attention, we here demonstrate a much simpler approach, convolutional in time and making use of (periodic) temporal encoding \cite{vaswani2017attention}.

The structure of this work is as follows: after reviewing related literature in \refSec{PreviousWork}, we introduce our approach \refSec{OurApproach}, involving data capture, and a convolutional GAN using time coordinate encodings.
Results are presented in \refSec{Results} before concluding in \refSec{Conclusion}.

\mysection{Previous Work}{PreviousWork}

There are several broad strategies that have been employed in order to solve the problem of generating novel motions from a limited set of motion captured examples. 
The first distinction is between kinematic and physics-based approaches. 
In kinematic approaches, plausible pose sequences/motion are generated without attention paid to physical properties. 
In physics-based approaches, a physical model is implemented and physics-based control signals (e.g. joint torques) are selected to create physically-plausible motion that responds to elements of the environment. 
Our approach belongs to the former category, so we start by reviewing works of this kind in this section.

\mysubsection{Non-deep learning approaches}{RWNonDeepLearning}

Within the category of kinematic approaches, a variety of techniques have been employed to generate motion examples. 
Older approaches involve stringing together and blending motion clips, like in \cite{safonova2007construction}, which uses a motion graph \cite{kovar2008motion,arikan2002interactive,lee2002interactive} constructed from mocap samples to generate new motion by interpolating two paths through the graph.
These two paths with a satisfactory interpolated motion are found through A* search.

Another way of blending motions is using kernel-based methods (e.g. \cite{rose1998verbs,levine2012continuous}) that learn an embedding of high-dimensional full-body motion in a low-dimensional latent space and can generate new, intermediate/blended motions by selecting points in the latent space.

A recent approach \cite{starke2019neural} uses shallow, 3-layer neural networks to learn multiple experts that output the next frame of motion given the previous frame (a motion prediction network), and learns an appropriate weighting of the experts based on the phase and goal of the motion (a gating network).
These networks are trained on an augmented motion capture dataset.

In recent years, however, deep learning techniques have shown promise in many applications,not only for motion denoising and motion reconstruction \cite{huang2018deep}, but also for motion generation. 
We review some recent advances in the next section.

\mysubsection{Deep learning generative approaches}{RWDeepLearning}

Recurrent neural networks (RNNs) are a type of deep neural network that specialize in generating sequential or time series data. 
Long short-term memory units (LSTMs) are an alteration that overcomes some of the vanilla RNN's shortcomings, like the vanishing gradient problem. 
These networks have shown promise at generating human motion sequences \cite{fragkiadaki2015recurrent, ghosh2017learning}.
RNNs or LSTMs can be combined with a generative adversarial network (GAN) structure, as \cite{ferstl2019multi} do to generate gestures.

While RNNs specialize in sequential data, there is some evidence that temporal convolution may be more effective than RNNs, at least for prediction involving long-term memory \cite{bai2018empirical}.
Similar to our work, \cite{holden2015learning} use temporal convolution, or convolutional neural networks, to learn from mocap examples and use it to generate motion sequences which they apply to mocap cleanup.
Their design is an autoencoder, typically introducing smoothing, which we prevent by making use of adversarial training.
As an aside, for phased motions, such as walking, another interesting approach is to rather than learn temporal information, is to instead provide the phase of the motion as input to the network, as shown by \cite{holden2017phase} for motion frame prediction.

Another possibility is to use normalizing flows, a technique similar to GANs that allows direct maximum likelihood estimation during training.
Henter and colleagues \cite{henter2019moglow} combine normalizing flows with LSTMs in order to generate human and animal locomotion.
This technique has also been applied to generating conversational gestures \cite{alexanderson2020style}.

\mysubsubsection{Extensions}{RWDeepLearningExtensions}

Techniques for learning patterns from motion examples can be combined with other techniques in order to generate more complicated sequences of motions. 
For example, Ling and colleagues \cite{ling2020mvaes} use Variational Autoencoders to learn a motion space, but combine it with reinforcement learning to generate motions to accomplish higher-level tasks.
\cite{holden2016deep} use the convolutional neural network from their 2015 work and use a feed-forward neural network to connect it with high-level control parameters like motion trajectories to give the user more control over the generated motion.

\mysubsection{Style transfer from video/images}{RWStyleTransfer}

Another promising avenue for generating novel motions is to use motion from non-mocap sources in order to drive or alter an existing motion capture clip or image.
Deep learning techniques have proven useful here as well. 
\cite{aberman2020unpaired} uses temporal convolution to learn motion information and combines it with time-invariant convolution to learn style information, which allows the means and variances of mocap motions like walking and jumping to be modulated by a style code derived from an unrelated video file. 
Instead of transferring video style to a motion capture motion, \cite{siarohin2019animating} use traditional convolutional techniques for processing images to transfer motion from a video to an input image, allowing a novel video to be created.

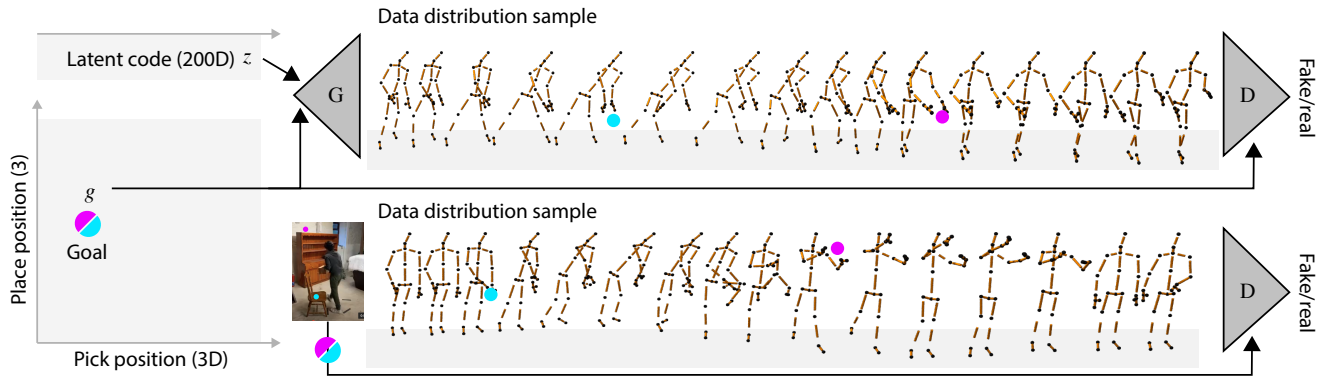
\mycfigure{Concept}{Our approach:
input to the generator $\generator$ is a latent code $\latentCoordinate$ and a pair of 3D vectors $\goal$ \textbf{(left)}.
Users can modify both vectors freely.
Output of the generator is a motion \textbf{(top)}.
This motion is fed into a discriminator $\discriminator$, shown \textbf{right}.
Learning maximizes confusion of that discriminator with respect to real and generated data \ie motions of the data distribution \textbf{(bottom)}.
}

\mysubsection{Physics-based approaches}{RWPhysics}

Physics-based approaches fall into two categories: approaches that incorporate motion capture as a reference for natural-looking motions, and pure physics-based approaches that do not make use of motion capture data.
In the first category are works by \cite{bergamin2019drecon} and  \cite{peng2018deepmimic}, both deep reinforcement learning (DRL) approaches that incorporate closeness to motion capture as part of the reward function.
The second category includes work by \cite{clegg2018learning} on generating dressing motions using physics simulation and deep reinforcement learning as well as the approach by \cite{yu2018symmetric} that generates locomotion using a DRL method that penalizes asymmetry in the loss function.

\mysection{Our Approach}{OurApproach}
We first explain how we represent the action goals (\refSec{InputRepresentation}) and the resulting human reach-and-place action (\refSec{OutputRepresentation}), how to acquire the data (\refSec{DataAcqusition}) before introducing our generative model (\refSec{Learning}).

\mysubsection{Input representation}{InputRepresentation}
We represent the motion constraints (\ie goals) as a sequence of 3D vectors.
As we are investigating reach-and-place motion only, this space is parameterized by a pair of 3D-vectors $\goal\in\mathbb R^{3\times 2}=\goals$.
The first vector qualifies where the object is to be picked up, the second where it is to be placed.

\mysubsection{Output representation}{OutputRepresentation}
We represent each reach-and-place action as a tuple $\action=\{\shape,\motion\}\in\actions$ of 3D shape $\shape\in\shapes$ and 4D motion $\motion\in\motions$ where $\actions,\shapes$ and $\motions$ are the spaces of action, shape and motion to be defined below.

Our aim is to design these spaces to be simple enough to define convolutions on them and ultimately apply a convolutional GAN.

\paragraph*{Shape}
We assume Kinect's standard topology.
The human shape  $\shape=\{\nodeDistance_i\}\in\shapes=\mathbb R^{\numberOfNodes}$ is represented by $\numberOfNodes=25$ nodes, a vector of distances in 3D world-space.
In other words, this is a stick figure, with each stick representing a defined distance between a pair of nodes, whilst also having the flexibility to change orientation arbitrarily.

\paragraph*{Motion}
We will make use of two interchangeable representations of motion: \emph{absolute} positional motion, and \emph{relative} orientational motion, both to be explained next.

First, absolute positional motion is a sequence of world space positions $\absoluteMotion=\{\nodePosition_{i,j}\}\in\mathbb R^{3\times\numberOfNodes\times\numberOfKeyframes}$ where every node $i$ has a position for each of the $\numberOfKeyframes$ animation frames $j$.
This is the form in which we will acquire data, how samples of the \emph{data distribution} look like and hence what the discriminator of a GAN will need to work with.

Second, relative motion is represented as time-varying direction $\relativeMotion = \{\nodeOrientation_{i,j}\}\in\mathbb{R}^{3\times\numberOfNodes\times\numberOfKeyframes}$, a sequence of direction vectors.
The orientation is relative to a coordinate frame defined from each node and its parent.
We first compute a matrix to align the $z$ axis with the direction from every node's positions at the first frame to its parent's position, also in the first frame.
The generator--the most relevant step of our approach--will work with this representation.
Note, that directions vectors are unit; multiplying them with the length of a bone will provide the position of a node.

The root node at time step 0 is considered to be at the origin in world-space.
Beyond time step 0, the root node is further represented as a time differential: the change of relative position per time step.

The procedure described above can be used to convert from absolute to relative motion and vice versa, denoted $\mathtt{aToR}(\absoluteMotion,\shape)$ and $\mathtt{rToA}(\relativeMotion,\shape)$.
Note that both steps are differentiable, without learned parameters.

\mysubsection{Data Acquisition}{DataAcqusition}
We directly capture our data distribution by acquiring 3D trajectories from a Kinect, recording at 30\,Hz. Within each sequence we record 25 marker, or node, positions representing the major joints of the human body.
In total, 600 sequences with durations of around 100 frames were captured.
We also acquire the goal positions.

All actions were performed by one male person, age 24, with no reported physiological conditions and on the same object, a box of 200\,g.
In this work, we do not generalize across subjects, kinds of goal objects, or any domain other than the domain of interest (goal start/end position).

All motion is recorded in world coordinates, forming the \emph{data distribution}, denoted $\dataDistribution$.
Samples from that distribution are noisy; a Kinect commonly makes mistakes in detecting pose.
For example, limbs may be recorded in entirely wrong orientations, and parts of the body jitter and jump.
While our results are not free from artifacts, we found those sampling artifacts to vanish, as in a denoising auto-encoder \cite{vincent2010stacked} but without the smoothing, thanks to our adversarial design.

We compute the shape $\shape$, \ie the node distances, from the Kinect scans directly as the average of the distances between a node and its parent across all takes and all frames.
Further, whitening is applied to the data by removing the intra-action mean and dividing by intra-action variance.

\mysubsection{Learning}{Learning}
We perform adversarial \cite{goodfellow2014generative}, \ie unsupervised, learning of a generator network to map from a latent coordinate and a pair of goals to a complete action, such that a discriminator network cannot tell apart samples produced by the generator and samples drawn from the data distribution.

More formally:
\begin{align*}
\underset{\learnableParameters}{\operatorname{argmin}}
\
\mathbb E_{\latentCoordinate\sim\latentSpace} 
\discriminator_\learnableParameters(\generator_\learnableParameters(\latentCoordinate|
\goal))-
&\mathbb E_{\dataSample\sim\dataDistribution(\goal)}
\discriminator_\learnableParameters(\mathtt{aToR}(\dataSample,\shape)|\goal)
,
\end{align*}
where $\mathbb E_{x\sim X}[\cdot]$ is the expected value operator sampling distribution $X$ with sample $x$,
$\discriminator$ is the discriminator to classify an action as belonging to the data distribution or not,
$\generator$ is the generator, mapping the latent code $\latentCoordinate$ from the latent space $\latentSpace$ to a complete action.
Both generator and discriminator share the tunable parameters $\learnableParameters$.
Rather than use the GAN loss as originally conceived, we instead make use of the Wasserstein loss \cite{arjovsky2017wasserstein}, which remedies the issue of vanishing gradients and allows for much more stable training.
 
\paragraph*{Discriminator}
Recall that the discriminator has to accept samples $\dataSample$ from both the data distribution $\dataDistribution$ as well as from the generator distribution $\generator(\latentCoordinate)$.
As we prefer to generate in relative space, we transform the data samples before showing them to the discriminator.

\begin{table}[h!]
  \begin{center}
    \caption{
    Discriminator architecture.
    }%
    \label{tbl:Discriminator}%
    \resizebox{\columnwidth}{!}{%
    \begin{tabular}{llrrr}
      Layer &
      &
      Feat. & 
      Size &
      Stride\\
      \toprule
      Position& $\absoluteMotion$, $\numberOfNodes\times 3=25\times3=$ &75& 32\\
      Goal& $\goal$, $2\times3=$ & 6 & 32\\
\midrule
      \texttt{concat} & & 81 & 32 &\\
      \texttt{coordCode}& $t$, $\sin(t)$, $\sin (2t)$ & 84 & 32\\
    \texttt{conv} & & 84 & 16 & 2\\
      \texttt{conv} & & 162 & 16 & 1\\
      \texttt{conv} & & 162 & 8 & 2\\
      \texttt{conv} & & 324 & 8 & 1\\
      \texttt{conv} & & 324 & 4 & 2\\
      \texttt{dense} & & 1200 & 1\\
      \texttt{dense} & & 1 & 1\\
      \bottomrule
    \end{tabular}%
    }
  \end{center}
\end{table}

The input to the discriminator $\discriminator(\dataSample|\goal)\in\mathbb \actions\rightarrow\mathbb R$ is an entire action $\dataSample$ and a goal $\goal$ condition.
Output is a single scalar, denoting the probability of the input action to belong to the data distribution.
\refTbl{Discriminator} gives the detailed architecture.

We only look at the absolute space-time-tensor $\absoluteMotion$ for motion. Hence, when using this discriminator, shape $\shape$ is just a time-constant set of values, informed by gradients, that together with a motion, can explain the data.
In theory, the generator might produce implausible shapes that then with implausible motion generate the data.
This however does not happen.
networks tend to find the simplest explanation (parsimony) which very often is also the truth: network parameters making up ``funny'' skeletons and even more ``funny'' motion might be harder to find, perform worse, or simply do not even exist, than the ones that actually cater to the inductive bias.
Using relative motion, hierarchical convolutions and quaternions, might be for human action, what the inductive bias of Deep Image Prior \cite{ulyanov2018deep} was for images.

This task is performed using a convolutional encoding of $\relativeMotion$ and $\goal$.
We start with a $(\numberOfNodes+(2\times3))\times\numberOfKeyframes$ tensor that stacks the time-varying orientation onto the two 3D goal vectors repeated over all keyframes.
At each level of this encoding, the input is first convolved with a bank of filters, second passing every filter result through a non-linearity.
The coefficients of the filter and the non-linearities (leaky ReLU) are the learned parameters $\learnableParameters$.
For image processing, these convolutions are two-dimensional, \eg $3\times 3$.
For a skeletal animation we use 1D convolutions in time instead, \ie $3\times 1$.
Note that convolution is translation-invariant, \ie the same input produces the same output, regardless of where it is in time.
This has benefits, but also drawbacks as explained next.

Additional to the action itself, we make use of \emph{coordinate encodings}.
Such an encoding adds additional features to the input of a convolutional neural network.
While classic convolution encodes \emph{what} a feature is, coordinate encodings hint to \emph{where} -- as we deal with time this means \emph{when}-- a feature is present.
In practice, this means a seemingly trivial change: at the first level, convolution will not only get the 3D positions, but also their time code.

A direct coordinate encoding has the drawback that it is very hard for a NN to capture high frequency response to that coordinate, first noted with respect to attention in language processing  \cite{vaswani2017attention}.
Hence it has been suggested, to also include periodic functions of coordinates, \ie not only $t$, but also $\sin(t)$, $\sin(2t)$ etc. in the feature vector.
We do so for two octaves.
The operator $\mathtt{coordCode}$ computes these periodic signals and a linear ramp and adds the three channels to the tensor.

Such a detector is strong as it can find evidence for being fake at all time positions (convolutionality), whilst it is also informed about absolute positions as well (positional encoding) and finally consolidates information across all scales (hierarchical).

\paragraph*{Generator}
The generator maps a latent coordinate $\latentCoordinate$ and the goal $\goal$ to a relative action representation $\relativeMotion$.
We mirror the design of the discriminator, a conditioned convolutional encoder, to a conditioned convolutional encoder-decoder.

Directly producing 3D-vectors can unfortunately produce direction vectors, that are not unit vectors.
We don't want to ask for the normalization from the network, neither do we want tolerate deviations from the desired stick-figure length. 
Hence, we simply normalize every direction vector before it is being used, a differentiable operation.
Note, that it is not sufficient to work with a network that generates only two degrees of freedom, \eg $x$ and $y$ and then solve for the third one, $z$, as there are two unit vectors $x,y,z$ and $x,y,-z$.

The generator also uses coordinate encodings both on the encoding and on the decoding step and on all levels.

For input we use a latent space of size 200, which during training is drawn from a normal distribution with $\mu$ of 1 and $\sigma$ of 100. We then simply concatenate the start and end goal coordinates (scaled between -1 and 1) to create an input vector of size 206. The first layer is a fully connected layer which projects the input vector of size 206, to a tensor of size $600\times4$. The subsequent upsampling and convolutional layers give an output equivalent in size to the discriminator input, (analogous to $\relativeMotion$) $75$$\times32$ time steps.

\refTbl{Generator} gives the detailed architecture of the network.

\begin{table}[h!]
  \begin{center}
    \caption{Generator architecture.}
    \label{tbl:Generator}
    \resizebox{\columnwidth}{!}{%
    \begin{tabular}{llrrr}
      Layer&
      &
      Feature&
      Size&
      Stride\\
      \toprule
      Goal & $\goal$ & 6\\
      Latent code  & $\latentCoordinate$ & 200\\
      \midrule
      \texttt{concat} & & 206 & &\\
      \texttt{dense} & & 600 & 4 &\\
      \texttt{coordCode} & $t, \sin(t), \sin(2t)$ & 607 & 4 &\\
      \texttt{upsample} & & 300 & 8 & 2\\
      \texttt{conv} & & 300 & 8 &1\\
      \texttt{upsample} & & 150 & 16 & 2\\
      \texttt{conv} &  & 150 & 16 &1\\
      \texttt{upsample} & & 75 & 32 & 2  \\
      \texttt{conv} & $\numberOfNodes\times 3=25\times3=$ & 75 & 32 & 1\\
        \bottomrule
    \end{tabular}%
    }
  \end{center}
\end{table}

\mysection{Results}{Results}

We report qualitative results as well as comparison to a range of ablations.

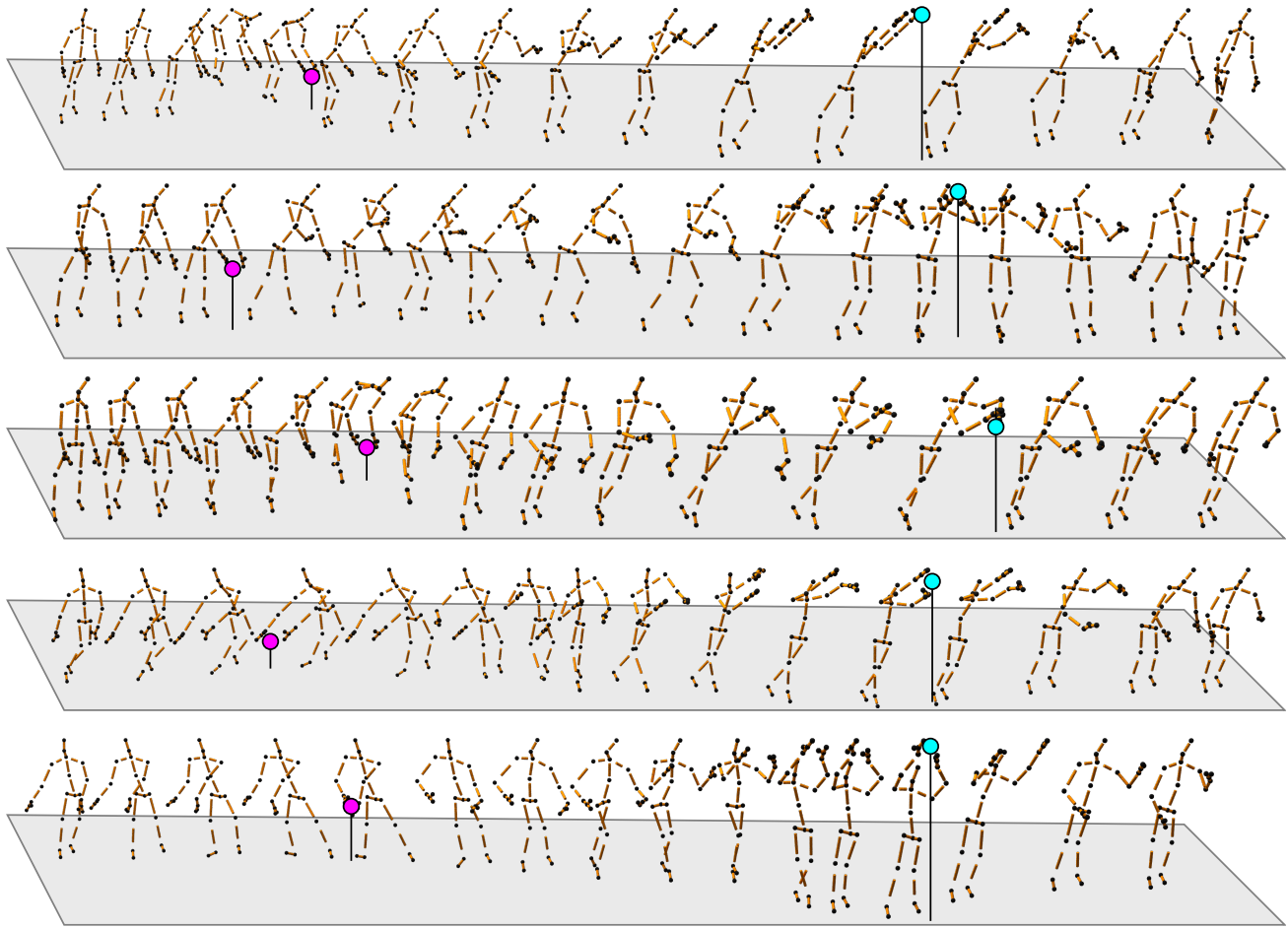
\mycfigure{Results}{Results of our method \textbf{(right)} for five different goal configurations
\textbf{(rows)} shown \textbf{(left)}.
The \textbf{first} action picks an object from the ground and the character stretches far up to place it.
After the place-action, a slightly unnatural, round, pose for the legs is noticeable.
Note how the character steps back to bend down.
In the \textbf{second} action, the character leans forward to place the object at head level.
In the \textbf{third} action, only one hand is used to hold the object.
The \textbf{fourth} action has a wide horizontal extend, resulting in a turning of the shoulder to reach left-down and place right-up.
In the \textbf{final} action, this is compared with a back-to-front motion, involving foot-stepping and some rich details like a horizontally tilted hip.
Note: the goal placement and motion is best seen from the supplemental video.}

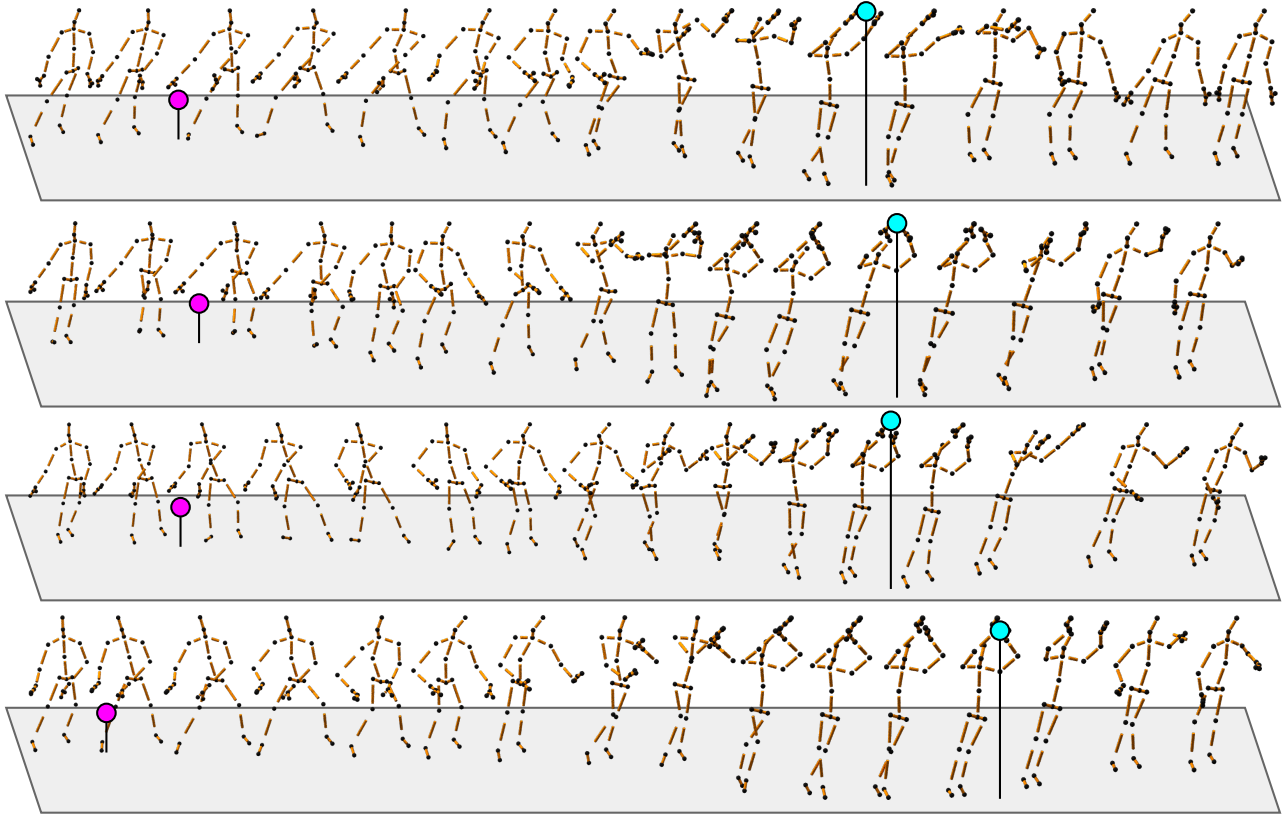
\mycfigure{Variation}{Provided the four same goals, our method samples from the distribution of  possible animations when provided different latent codes.
Despite the same goals: note the entirely different foot-stepping in each.
In sample one and two the character is upright, while in three and four the back is bent.
In all samples, the head is ``leading'' the action.
Also note how the time-point at which the goal is reached, differs, as we only prescribe it to be reached, not when.
We here visualize the goal only at the frame where the hands are closest, the goal's location however is the same in the 4 samples.}

\paragraph*{Qualitative results}
Qualitative results are seen in \refFig{Results}.
We see that the start and end pose are faithful human poses and the in between ones are plausible.
They reach both goals in a reasonable posture and blend between them for all body parts.
The weakest aspect are the feet: while upper extremities move plausible, the feet can sometimes appear to slide on the ground.
Note, how the results are not perfect, but non-noisy, while samples from the data distribution have noise, \eg for some frames, it can happen that a leg or arm is flipped, indicating the system acted as a denoising generative model.

\refFig{Variation} shows the distribution of results when the goals are held fixed and different samples are drawn by changing the latent code.
Despite the goals being constant, the animation however is different, while each is still a plausible motion reaching the goals, but in different ways.

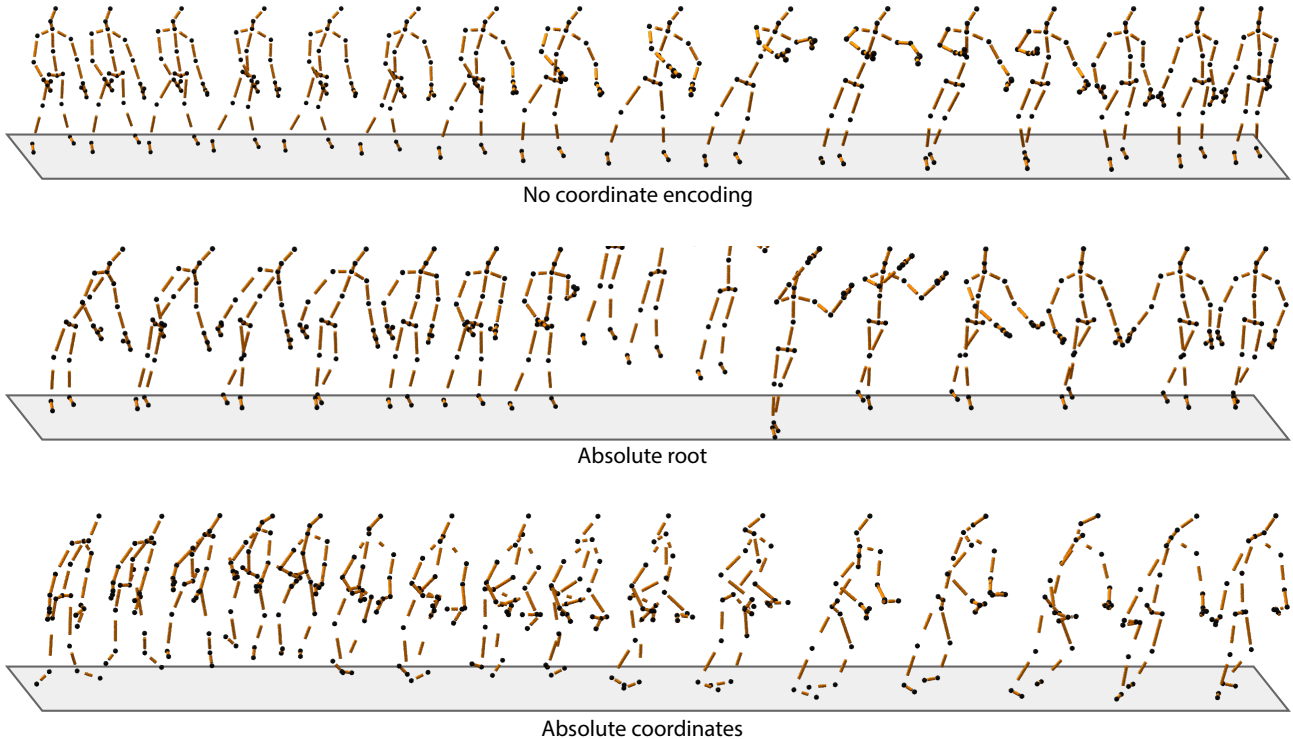
\mycfigure{Ablation}{Three ablations of our approach, from top to bottom.
The \textbf{top} one is the full method, but without coordinate encoding, \ie the convolutions do not have access to the time code or a nonlinear, periodic function of it.
The result has from-the-book low-pass artefacts, best seen also in the video: the feed are sliding over the ground, everything looks plausible, just low-passed as the NN has no access to high-frequency information.
The \textbf{middle} one is a version where the root node position is encoded relative to the world center \ie absolute at every time frame.
Here the entire skeleton drifts.
The \textbf{bottom} performance results from using absolute coordinates instead of relative ones.
We see that the figure is non-rigid, indicating the discriminator does not provide gradients, or it is simply very hard to generate results that fulfill constant-distance constraints.
Our approach has these built-in, while still being generative.
Again, all motions for all ablations are independent random samples of random goals and not meant to be compared to each other but judged on their own, no-reference.}

\paragraph*{Ablations}
\refFig{Ablation} studies three ablations focusing on the main contribution that allows using convolutions from motion.

The first ablation (\refFig{Ablation}, top) is to not use coordinate encodings.
We find that the skeleton has rigid bones, but the motion itself is subject to low and mid-frequency noise.
The most salient mistake is to not make distinct gestures like footsteps, but everything occurs in a temporally-smooth underwater motion.
This probably is, as the NN has to make up functions from boundary effects that support temporal locality.
Using coordinate encodings, this effect is drastically reduced, as again best seen in the supplemental video.

The second ablation is to handle the root node like any other node: relative in space, but not relative in time
(\refFig{Ablation}, middle).
This results in a patchwork of gestures that are plausible per-se but do not fit together.

The third ablation (\refFig{Ablation}, bottom) is to use absolute 3D positions throughout.
This makes the implementation much simpler and for our exposition one may simply assume $\mathtt{aTor}$ and $\mathtt{rToa}$ to be identity.
We find, that samples of this representation lack rigidity: the distance along bones is free to change.
The NN does a reasonable job in maintaining a rigid ensemble stick-figure, but that comes at the expense of adhering to a constraint that is easy to enforce by-design.

From these ablations, which are best also seen repeated in the supplemental video, we conclude the representation, discriminator and generator are good examples of learning with hard-coded physical constraints.

\mysection{Conclusion}{Conclusion}
We have presented a generative model of reach-and-place motion.
Starting from human motion labeled by goals, the model, provided new goals, produces new motion.

We have reported qualitative results, demonstrating several aspects of our approach are superior to important ablations.
Regrettably, as with many generative models, quantifying the success is difficult: we cannot simply use an L2 between any reference and what we generate, while existing measures of diversity, such as FID score are relative to image-specific properties.
Accordingly, we have shown samples from the generated and data distribution ---to be judged on their own, not in respect to a reference--- in conjunction with ablations.

Present circumstances dictate that it is quite difficult to capture large amounts of motion capture data, however post pandemic it will be possible to capture data from more subjects and motions to generalize the model further.
In future work, we would like to extend to different forms of goals, different subjects and objects to move, as well as use the model in other applications such as motion denoising, or motion in-painting. 

{\small
\bibliographystyle{ieee_fullname}
\bibliography{paper}
}

\end{document}

%% file: tr-commands.tex
\usepackage{graphicx}
\usepackage{soul}
\usepackage{amsmath}
\usepackage{amssymb}
\usepackage{microtype}
\usepackage{wrapfig}
\usepackage{xcolor}
\usepackage{booktabs}
\usepackage{multirow}

\def\figurePath{Figures/}

\def\mycfigure#1#2{%
    \begin{figure*}[htb]%
    \centering\includegraphics*[width = \linewidth]{\figurePath#1}%
    \caption{#2}%
    \label{fig:#1}%
    \end{figure*}%
}

\newcommand{\refSec}[1]{Sec.~\ref{sec:#1}}
\newcommand{\refFig}[1]{Fig.~\ref{fig:#1}}

\newcommand{\refTbl}[1]{Tbl.~\ref{tbl:#1}}

\soulregister\ref7
\soulregister\cite7
\soulregister\refFig7
\soulregister\cite7
\soulregister\ref7
\soulregister\pageref7
\soulregister\shortcite7
\soulregister\eg0
\soulregister\ie0
\soulregister\etal0

\DeclareGraphicsExtensions{.png,.jpg,.pdf,.ai,.psd}
\newcommand{\mysection}[2]{\section{#1}\label{sec:#2}}
\newcommand{\mysubsection}[2]{\subsection{#1}\label{sec:#2}}
\newcommand{\mysubsubsection}[2]{\subsubsection{#1}\label{sec:#2}}